# Effect of Hyper-Parameter Optimization on the Deep Learning Model Proposed for Distributed Attack Detection in Internet of Things Environment

Md Mohaimenuzzaman[1,*], Zahraa Said Abdallah[1], Joarder Kamruzzaman[2], Bala Srinivasan[1]
1. Faculty of Information Technology, Monash University, Australia
2. Faculty of Science and Technology, Federation University, Australia

## ABSTRACT

*This paper studies the effect of various hyper-parameters and their selection for the best performance of the deep learning model proposed in [1] for distributed attack detection in the Internet of Things (IoT). The findings show that there are three hyper-parameters that have more influence on the best performance achieved by the model. As a consequence, this study shows that the model's accuracy as reported in the paper is not achievable, based on the best selections of parameters, which is also supported by another recent publication [2].*

## INTRODUCTION

Diro and Chilamkurti [1] have introduced a distributed deep neural network model for intrusion detection in the IoT environment. The primary principle behind the proposed model is to train the deep neural network model using multiple nodes in a distributed computing environment while parameter sharing and optimization is done through a coordinating master node. The proposed approach is of interest because of the reported best known accuracy on the NSL-KDD benchmark data set. For example, for 2-class and 4-class NSL-KDD benchmark datasets, it is reported that their model archive the test accuracy of 99.20% and 98.27% respectively. However, another very recent article by Yin et al. [2] provides a comprehensive evaluation of the performances of a wide range of learning models including their proposed deep leaning model on the same NSL-KDD benchmark dataset. Their best reported accuracies are 83.28% and 81.29% respectively for the above two types of classifications. We aimed to replicate both the studies for validating the results. Studies by Yin have been reproduced and validated with the reported accuracy based on their published parameter specifications. However, with the absence of the source code and chosen hyper-parameter values for the reproducibility of reported accuracies in [1], we investigated the effect of hyper-parameters and their inter-relationship to come up with the best model configurations for best performance. Surprisingly, the end result is contradictory to that of the reported results.

**EXPERIMENTAL SETUP**

In order for the comparison to be meaningful and fair, we adopted the same pre-processing steps, network architecture, parameter and hyper-parameter values that were explicitly mentioned in [1]. The application of the 1-to-n encoding technique on the categorical features provides 122 input features in total. The label is converted into 2-class (Normal vs Attack) and 4-class (Normal, DoS, Probe and R2L.U2R). Finally, before training, the feature values are normalized using z-score normalization technique. According to the paper, the neural network models are built using the Keras deep learning API with Theano backend. The layers of the deep models are fully connected sequential layers (ANN) having 122 input features, 150 first layer neurons, 120 second layer neurons, 50 third layer neurons and a softmax layer as the output layer having neurons equals to the number of output classes, either 2 or 4. Although the paper mentions the use of hyper-parameters such as activation function, loss function, optimizer, mini-batch, epochs, learning rate and dropout rates, a number of their individual values are not explicitly mentioned for the accuracy reported in the paper. Hence, we varied the values of these hyper-parameters and their effects on the accuracy is presented in the next section.

**EFFECT OF HYPER PARAMETERS ON ACCURACY**

Among the listed hyper-parameters, the authors [1] mentioned that *mean squared error* is used as the loss function, *SGD* (Stochastic Gradient Descent) with constant learning rate is used as the optimiser, *softmax* is used as the activation function for the output layer and the number of iterations for *epoch* is set to 50. Furthermore, a list of activation functions (i.e. *tanh, rectified linear unit and maxout*) are mentioned for input and hidden layers, but it is not specified which one produces the highest accuracy. We chose the *rectified linear unit (ReLU)* as it is the mostly used one and since *maxout* is no more available in the latest Keras library [3]. Another set of missing information is the values of *dropout rate*, *batch size* and *learning rate* to achieve best results. Given the number of epochs is fixed, *learning rate* is important. In order to determine the best values of these hyper-parameters for both 2-class and 4-class deep learning models, we conducted experiments by varying the learning rates, batch sizes and dropout rates while keeping the same values of those hyper-parameters that are explicitly mentioned in the paper. In all our experiments, the batch size is fixed to 8, 16, 32, 48 and 64, and the models are trained on different combinations of dropout rates and learning rates.

Figures 1 and 2 show the accuracy of the 2-class and 4-class models with respect to different leaning rates under various combinations of dropout ratio and batch sizes. The legends for dropout percentage is displayed as *Dropout: x% -y%* where x% is the percentage of dropout applied in input layer and y% is the percentage of dropout applied to hidden layers. The last line graph (e) of both the figures shows the accuracy for different batch sizes while keeping the learning rate (lr) and dropout rate fixed for which the best accuracy achieved for that particular type of classification model.

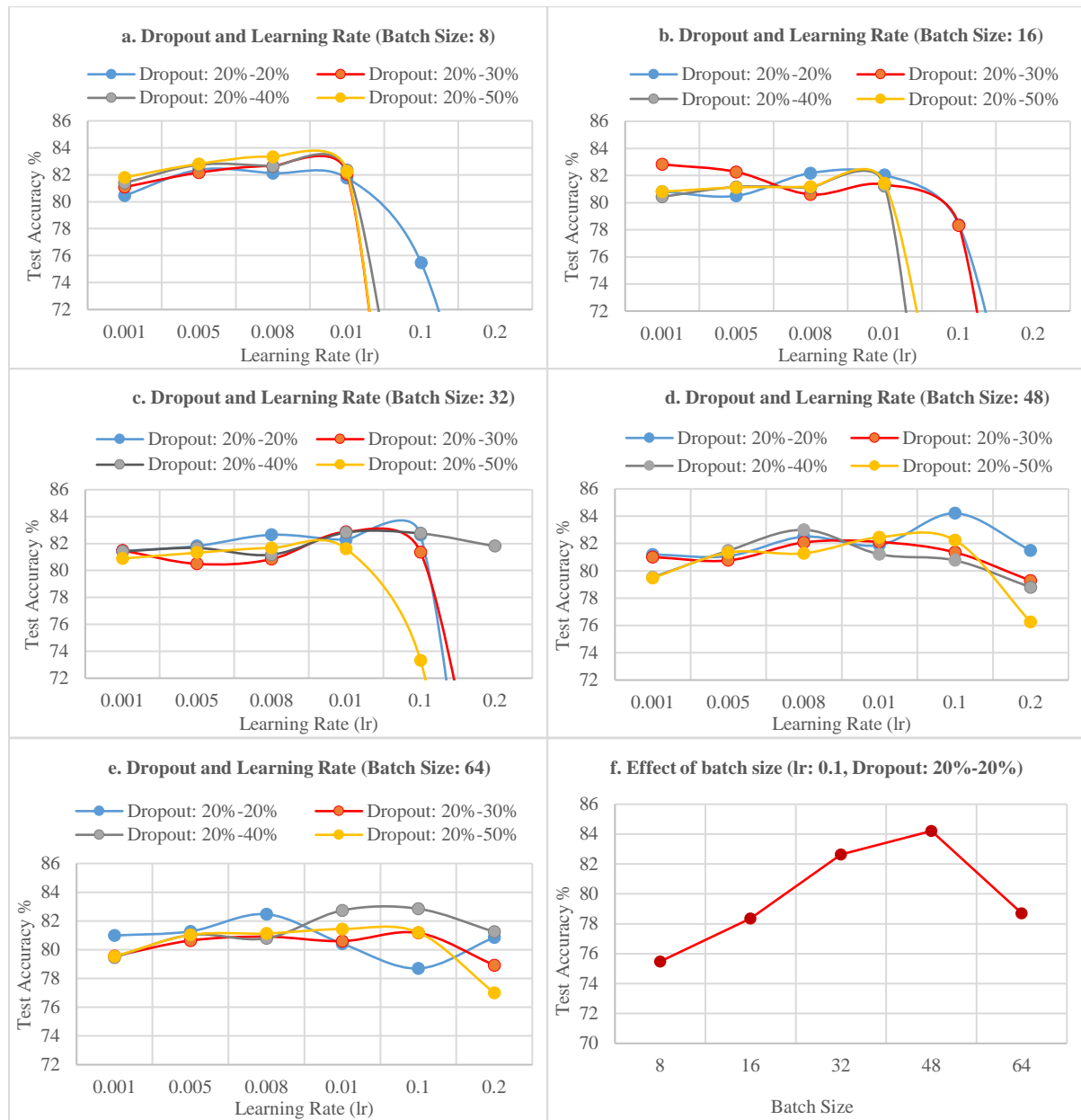

Fig 1: Test accuracies by 2-class classification model during hyper-parameter optimization

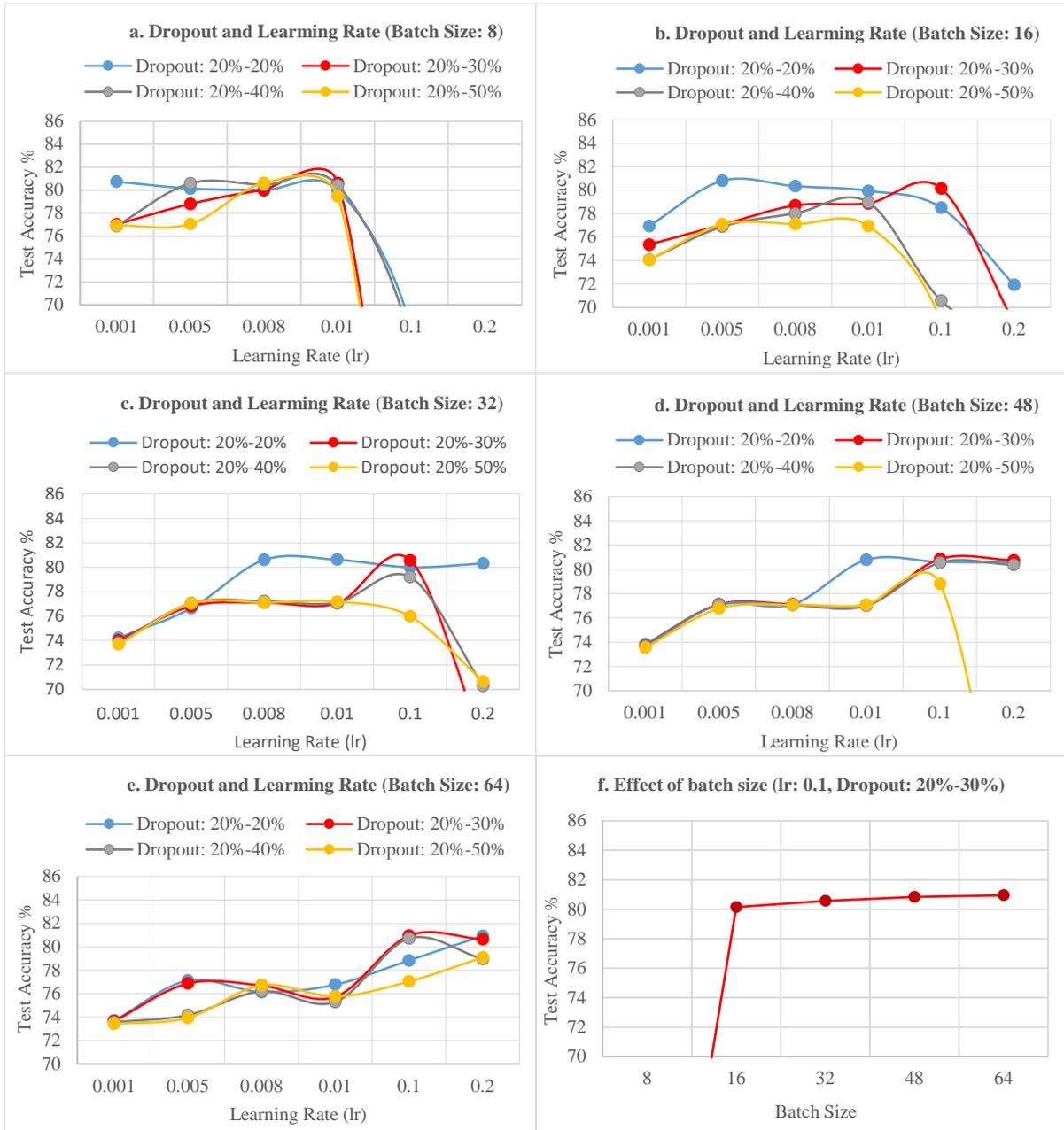

Fig 2: Test accuracies by 4-class classification model during hyper-parameter optimization

One general observation that can be inferred from these sets of graphs (in Figs. 1 and 2) is that the best accuracy that can be achieved for the 2-class data set is 84.21% with a learning rate of 0.1, batch size of 48 and dropout rate of 20% for input layer and 20% for hidden layers. Similarly, for the 4-class data set, the best accuracy is 80.95% when the learning rate is 0.1, batch size is 64 and dropout rate is 20% for input layer and 30% for hidden layers. Table 1 provides the model configuration (combination of hyper-parameters) for both the classes under the best accuracy scenario. If we look at that table, we see that the *kernel_initializer* is set to 'normal' which makes sure that the initial random weights are normally distributed. Furthermore, the values for *kernel_constraint* and *momentum* are chosen by following the original research article [4] that proposed the idea of the application of dropout. The values for the parameters marked with asterisk (*) are not specified in the paper [1] and are chosen by us after our rigorous hyper-parameter optimization. Almost identical parameter values achieve the best accuracy for both the classes.

```
Configuration: 2-Class                  Configuration: 4-class
No. of INPUTS: 122                      No. of INPUTS: 122
No. of Neurons:                         No. of Neurons:
   First Hidden Layer: 150                 First Hidden Layer: 150
   Second Hidden Layer: 120                Second Hidden Layer: 120
   Third Hidden Layer: 50                  Third Hidden Layer: 50
   Output Layer: 2                         Output Layer: 4
Activation Function                     Activation Function
   Visible and Hidden layers: relu         Visible and Hidden layers: relu
   Output Layer: softmax                   Output Layer: softmax
Loss Func: mean_squared_error           Loss Func: mean_squared_error
Epochs: 50                              Epochs: 50
*Batch Size: 48                         *Batch Size: 64
*Learning Rate: 0.1                     *Learning Rate: 0.1
*Dropout in input layer: 20%            *Dropout in input layer: 20%
*Droput in hidden layers: 20%           *Droput in hidden layers: 30%
*kernel_initializer: normal             *kernel_initializer: normal
*kernel_constraint: maxnorm(3)          *kernel_constraint: maxnorm(3)
*Decay: 0                               *Decay: 0
*Momentum: 0.9                          *Momentum: 0.9
Optimization: SGD                       Optimization: SGD
```

*values are chosen through our hyper parameter optimization process

Table 1: Deep model configuration achieved after hyper-parameter optimization.

**COMPARATIVE ANALYSIS**

Table 2 summaries the accuracy as reported by [1], [2] and our above experimental evaluation. From this table, it is noticeable that there is a big jump in the test accuracy reported by Diro and Chilamkurti [1] regardless of classification and model types. On the other hand, the accuracies found in our implementation are close to the accuracies reported by Yin et al. Interestingly, the reported test accuracy by Diro and Chilamkurti is very similar to the training accuracy reported by Yin et al. (99.81% for 2-class and 99.53% for 5-class respectively) and to the training accuracies of our implemented model (99.00% for 2-class and 98.46% for 4-class respectively) as well. These observations made us to believe that the reported accuracy in [1] might have actually misreported as the test accuracy instead of training.

|  | Diro and Chilamkurti [1] | | Yin et al. [2] | | Our Study on [1] | |
|---|---|---|---|---|---|---|
| Classification Type | 2-Class | 4-Class | 2-Class | 5-Class | 2-Class | 4-Class |
| Shallow Model | 95.22 % | 96.75 % | - | - | 80.39 % | 78.68 % |
| Deep Model | 99.20 % | 98.27 % | 83.28 % | 81.29 % | 84.21 % | 80.95 % |

Table 2: Test accuracies reported in different articles and from our implementation of [1]

# CONCLUSION

Our extensive experimental evaluation demonstrates that the level of accuracy reported in [1] is not reproducible using the proposed method and hence, in fact, has overstated the method's capability. We were able to achieve the reported level of accuracy only with the training data which is basically the model's training accuracy, not the model's test accuracy. Since this is a significant research area attracting a large number of researchers around the globe, the true capability of the proposed method needs to be assessed and reported accurately in order to avoid misleading the researchers.